\pdfoutput=1

\documentclass[11pt]{article}

\usepackage[preprint]{acl}

\usepackage{times}
\usepackage{latexsym}
\usepackage{tabularx}
\usepackage{graphicx}

\usepackage{graphicx}                    
\usepackage{amsmath}                     
\usepackage{amsfonts}                    
\usepackage{amssymb}                     
\usepackage{amsthm}                      
\usepackage[normalem]{ulem}              
\usepackage{setspace}                    
\usepackage{xcolor}
\usepackage{multirow}
\usepackage{amsmath}
\usepackage{amssymb}
\usepackage{booktabs}
\usepackage{times}
\usepackage{epsfig}
\usepackage{graphicx}
\usepackage{amsmath}
\usepackage{amssymb}
\usepackage{tabularx,booktabs}
\usepackage{bm}
\usepackage{icomma}

\usepackage[T1]{fontenc}

\usepackage[utf8]{inputenc}

\usepackage{microtype}

\usepackage{enumitem}

\usepackage{inconsolata}
\usepackage{hyperref}
\usepackage[capitalize]{cleveref}
\crefname{section}{Sec.}{Secs.}
\Crefname{section}{Section}{Sections}
\Crefname{table}{Table}{Tables}
\crefname{table}{Tab.}{Tabs.}
\usepackage{changes}

\usepackage{times}
\usepackage{latexsym}
\usepackage{tabularx}
\usepackage{graphicx}                    
\usepackage{amsmath}                     
\usepackage{amsfonts}                    
\usepackage{amssymb}                     
\usepackage{amsthm}                      
\usepackage[normalem]{ulem}              
\usepackage{setspace}                    

\usepackage{amsmath}
\usepackage{amssymb}
\usepackage{booktabs}
\usepackage{times}
\usepackage{epsfig}
\usepackage{graphicx}
\usepackage{amsmath}
\usepackage{amssymb}
\usepackage{tabularx,booktabs}
\usepackage{bm}
\usepackage{graphicx}
\usepackage{colortbl}

\usepackage[T1]{fontenc}

\usepackage[utf8]{inputenc}

\usepackage{microtype}

\usepackage{inconsolata}

\usepackage[normalem]{ulem}

\newcommand{\commentOUT}[1]{}
\title{Gloss2Text: Sign Language Gloss translation using LLMs and Semantically Aware Label Smoothing}

\author{Pooya Fayyazsanavi \\
  George Mason University \\
  \texttt{pfayyazs@gmu.edu} \\\And
  Antonios Anastasopoulos \\
  George Mason University \\
  \texttt{antonis@gmu.edu} \\\And
  Jana Košecká \\
  George Mason University \\
  \texttt{kosecka@gmu.edu} \\}

\begin{document}
\maketitle
\begin{abstract}
Sign language translation from video to spoken text presents unique challenges owing to the distinct grammar, expression nuances, and high variation of visual appearance across different speakers and contexts. 
The intermediate gloss annotations of videos aim to guide the translation process. In our work, we focus on {\em Gloss2Text} translation stage
and propose several advances by leveraging pre-trained large language models (LLMs), data augmentation, and novel label-smoothing loss function exploiting gloss translation ambiguities improving significantly the performance of state-of-the-art approaches. 
Through extensive experiments and ablation studies on the PHOENIX Weather 2014T dataset, our approach surpasses state-of-the-art performance in {\em Gloss2Text} translation, indicating its efficacy in addressing sign language translation and suggesting promising avenues for future research and development.\footnote{Code can be found at the following GitHub repository: \href{https://github.com/pooyafayyaz/Gloss2Text/tree/main}{https://github.com/pooyafayyaz/Gloss2Text/tree/main}}.
\end{abstract}

\section{Introduction}





\begin{figure}[t]
\centerline{\includegraphics[width=0.9\linewidth]{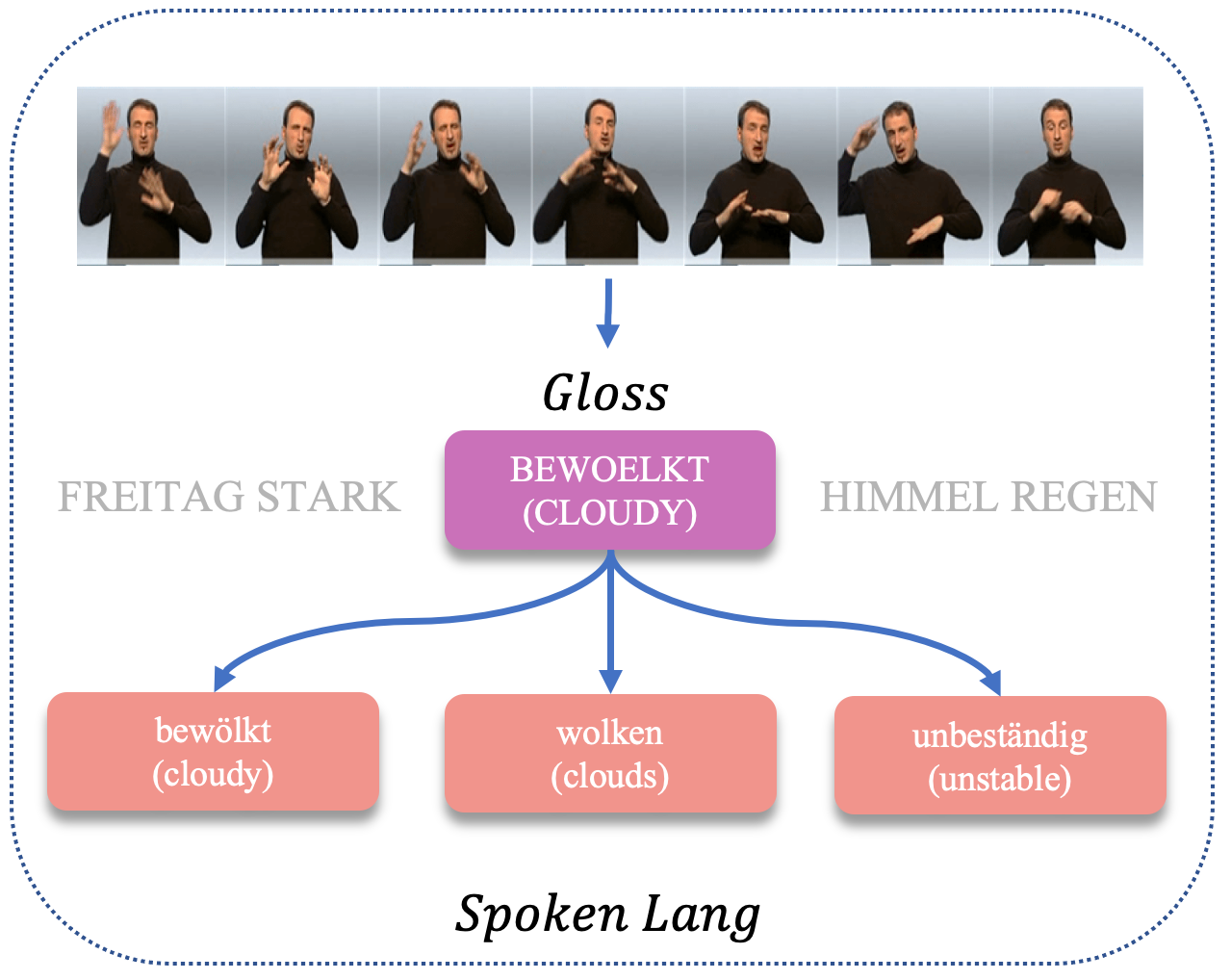}}
\caption{An example of ambiguity in sign language is demonstrated by the gloss "BEWOELKT (CLOUDY)," which is represented in multiple translations within the dataset. As shown, ambiguity may share the same meaning but differ in form, such as "wolken (cloudy)," or where the gloss represents the concept meaning, such as "unbeständig (unstable)."}
\label{fig:motiv}
\vspace{-1em}
\end{figure}

Sign language translation from video to spoken text often involves two phases: {\em Sign2Gloss} and {\em Gloss2Text}. In {\em Sign2Gloss} phase, the gloss annotations, are predicted from input videos as shown in the top part of Figure \ref{fig:motiv}, establishing a link between visual expressions and corresponding meanings. The subsequent {\em Gloss2Text} phase, translates these gloss annotations into spoken language. While gloss annotations have strong limitations as a linguistic representations~\citet{angelova2022using}, the emergence of pre-trained large language models, word embeddings, and advances in Machine Translation open new possibilities for improvements in {\em Gloss2Text} translation task. 
%
In our work, we propose to leverage Large Language Models (LLMs) pre-trained on expansive and diverse corpora along with novel sign language specific label smoothing loss and data augmentation techniques to improve {\em Gloss2Text} phase
of sign language translation task. 
\commentOUT{
Several recent approaches~\citet{ye2023scaling,chen2022two,zhu2023neural} that incorporate LLMs often involve fine-tuning the entire model. Still, with LLMs having millions or even billions of parameters, conventional fine-tuning may prove less effective due to potential overfitting. To overcome these challenges, we propose an efficient fine-tuning method using Low-Rank Adaptation (LoRA) \citet{hu2021lora}, enabling the utilization of larger models without encountering overfitting issues. Moreover, we utilize back-translation and forward-translation to mitigate the scarcity of sign language datasets by procuring a more diverse training set. Finally, as shown in Figure \ref{fig:motiv}, recognizing the ambiguities in gloss where identical gloss may yield different yet equivalent translated meanings, we propose to utilize a label smoothing where the incorrect predictions that are semantically similar are penalize less.
}
Our contributions are in:
\begin{itemize}[noitemsep,nolistsep,leftmargin=*]
\item Development of tailored data augmentation techniques for {\em Gloss2Text} translation, including paraphrasing to enhance spoken aspects by proxy language translation, and back-translation for gloss augmentation.
\item Novel label-smoothing loss function optimized for gloss translation specific ambiguities, reducing penalties for incorrect predictions that are similar to the target translation.
\item State-of-the-art performance in {\em Gloss2Text} translation, surpassing existing benchmarks on the PHOENIX Weather 2014T dataset 
and detailed ablation study of different components of our approach. 
\end{itemize}

\section{Approach}
The goal of our gloss translation system is to convert a series of gloss annotations $G = {g_1, g_2, \ldots, g_T}$ into a spoken word sequence $T = {t_1, t_2, \ldots, t_L}$. 
given $n$ pairs where each pair can have different input and output lengths. Our approach involves fine-tuning large language models tailored specifically for our task.


To effectively train a typical Neural Machine Translation (NMT) model, a corpus of around 1 million parallel samples is often required~\citet{sennrich2019revisiting}. However, the existing sign language datasets are of several orders of magnitude smaller. For instance, the PHOENIX-2014T German sign language dataset \citet{camgoz2018neural}, the most widely benchmark for continuous sign language, has only $8,257$ gloss-text pairs.

One group of approaches \citet{chen2022simple,chen2022two,zhou2023gloss} has concentrated on fine-tuning LLMs for the sign language {\em Gloss2txt} translation task without data augmentation. 
A series of studies~\citet{ye2023scaling,zhu2023neural,angelova2022using,zhang2021approaching} have investigated limited data augmentation techniques to address the challenge of data scarcity.

\begin{figure}[t]
\centerline{\includegraphics[width=\linewidth]{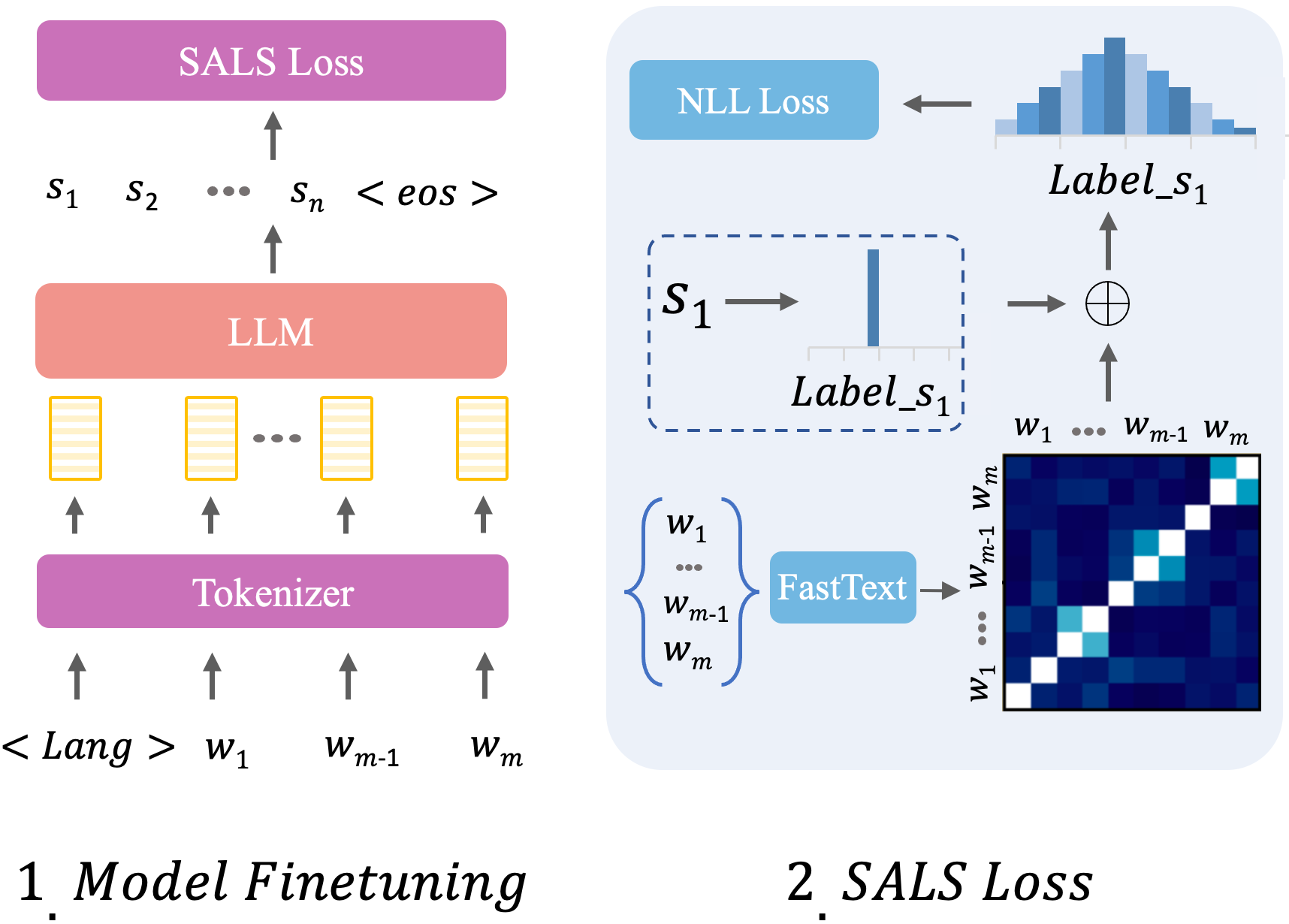}}
\caption{The proposed architecture for {\bf \em Gloss2Text} translation. Initially, the similarity of each word to others is compared. During training with label smoothing, depicted on the left side, the model aims to identify the most similar words to the target word and assign heightened labels to those words.}
\label{prposed_approach}
\vspace{-1em}
\end{figure}

\commentOUT{To address the scarcity of labeled data, we employ various data augmentation techniques on both gloss and translation sides. Additionally, we propose a novel label smoothing loss function to enhance the effectiveness of translation based on sign language characteristics. Figure \ref{prposed_approach} illustrates the overall architecture. In the subsequent sections, we will provide a detailed description of each component.
}

\setlength{\tabcolsep}{0.3\tabcolsep}
\begin{table*}[t]
\footnotesize
{\smallskip}

\begin{tabularx}{\textwidth}{l cccccc cccccc}
\hline
\multirow{3}{*}{Set} & \multicolumn{6}{c}{ Dev } & \multicolumn{6}{c}{ Test } \\
 & \multicolumn{4}{c}{ BLEU } & \multirow{2}{*}{ ROUGE } & \multirow{2}{*}{ CHRF++ }  & \multicolumn{4}{c}{ BLEU } & \multirow{2}{*}{ ROUGE } &  \multirow{2}{*}{ CHRF++ }\\
\cmidrule{2-5}
\cmidrule{8-11}
& 1 & 2 & 3 & 4 &  & & 1 & 2 & 3 & 4 &  &   \\


\midrule
1. \citet{camgoz2018neural}  & 44.40 & 31.83 & 24.61 & 20.16 & -- & -- & 44.13 & 31.47 & 23.89 & 19.26 & -- & -- \\
2. \citet{camgoz2020sign} & 50.69 & 38.16 & 30.53 & 25.35 &-- & -- &  48.90 & 36.88 & 29.45 & 24.54 & -- & --  \\
3. \citet{yin2020better} &  49.05 & 36.20 & 28.53 & 23.52 &-- & -- &  47.69 & 35.52 & 28.17 & 23.32 & -- & --  \\
4. \citet{chen2022two} & 53.57 & 40.18 & 31.93 & 26.40  & 52.50 & 49.55 & 52.81 & 39.99 & 31.96 & 26.43 & 51.66 & 49.76  \\
5. \citet{ye2023scaling} &  48.68 & 37.94 & 30.58 & 25.56 & -- & -- & 48.30 & 37.59 & 30.32 & 25.54 & -- & --   \\



6. NLLB-Zero Shot & 12.26 & 3.19 & 1.29 & 0.64 &  18.79 & 19.25 & 12.71 & 4.08 & 1.79 & 0.84 & 19.14 & 19.86  \\
7. NLLB-FineTuned   & 53.64 & 40.56 & 32.35 & 26.78  &  53.84 & 49.53 & 52.89 & 40.12 & 32.03 & 26.50 & 53.46 & 49.65  \\
8. NLLB-Aug   & 55.12 & 41.74 & 33.40 & 27.76  &  55.13 & 50.72  & 53.63 & 40.79 & 32.68 & 27.13 & 54.04 & 50.41  \\
9. NLLB-SALSloss & 55.22 & 42.04 & 33.56 & 28.05  &  55.26 & 50.65 &  53.26 & 40.92 & 33.00 & 27.55 & 54.28 & 50.01 \\

10. NLLB-all  &  $\mathbf{55.61}$ & $\mathbf{42.10}$ & $\mathbf{33.71}$ & $\mathbf{28.11}$   & 55.03 & $\mathbf{50.64}$ & $\mathbf{54.79}$ & $\mathbf{41.90}$ & $\mathbf{33.77}$ & $\mathbf{28.20}$ & $\mathbf{54.44}$ & $\mathbf{50.79}$ \\
\bottomrule
\end{tabularx}
\vspace{-1em}
\caption{Comparison with state-of-the-art methods on the PHOENIX-2014T dataset demonstrates the effectiveness of our framework, achieving higher performance despite having approximately a tenth fewer parameters.}
\label{tab:res}
\vspace{-1em}
\end{table*}



In our experiments, we propose several strategies for fine-tuning  
NLLB-200 \citet{costa2022no} model. Additional experiments with
alternative models including MT5 \citet{xue2020mt5}, mBart \citet{liu2020multilingual} can be found in Appendix \ref{sec:aptype}. These models share multilingual characteristics, enabling them to handle diverse language pairs efficiently. However, they differ in the datasets they are pre-trained on, the specific architectures they employ, and their respective training objectives, which affects their performance in sign language translation tasks.

\subsection{Data Augmentation}
To improve the robustness of the baseline translation approach, we explore two distinct data augmentation techniques. 

\paragraph{Paraphrasing}
translates the original target sentence into a proxy language (English) and then back to the original (German). This cycle introduces linguistic diversity on the target side while ideally preserving the original meaning in gloss annotations, exposing our model to a broader spectrum of linguistic variations.


\paragraph{Back Translation}
involves training a reverse translation model with the spoken data as input, producing the corresponding gloss sequence. If the model's generated gloss sequence differs from the original one, we iterate over sentences and incorporate this new gloss sequence as a silver label alongside the translation pair for our primary training process.

\subsection{Semantically Aware Label Smoothing}
In the conventional label smoothing approach~\cite{szegedy2016rethinking, muller2019does} one replaces  one-hot encoded label vector ${\bf y}_{hot}$ with a mixture of ${\bf y}_{hot}$ and the uniform distribution ${\boldsymbol{y_s}=(1-\beta) \cdot {\boldsymbol{y_{hot}}}+ \frac{\beta}{N}}$, where $\beta$ is a smoothing parameter.
%
With this approach however, 
probabilities for all words in the vocabulary are non-zero, including those not present in our target vocabulary.
\noindent
We propose a new vector of probabilities ${\bf y}_{sals}$ where for each word we first set the value of non-target words to zero. Among the words in the target vocabulary $V_{target}$, we compute the semantic similarity of each word $\{v_i\}_{i=1}^N$ with other words in the target vocabulary. 
We use FastText \citet{joulin2016fasttext} to generate word embeddings $\{w_i\}_1^N$ vectors for each word $v_i$ and then compute their cosine similarity. Therefore, we calculate the similarity values as follows:

\noindent The final semantically-aware vector of probabilities for word $v_i$, ${\bf y}^i_{sals}$ vector of probabilities  will be:
\begin{equation*}
{\bf y}_{sals}= \begin{cases}f_{sim}(w_i,w_j) &  \geq \lambda \wedge \forall v_j \in V_{target} \\ 
\frac{\beta}{N} &  < \lambda \wedge \forall v_j \in V_{target} \\ 
0 & \text {otherwise }\end{cases}
\end{equation*}
Three scenarios can occur: Firstly, for words in the target language with high similarity, defined as $\lambda$, we utilize the cosine similarity of their word embeddings, denoted as $f_{sim}(w_i,w_j) =  \frac{{w_i}^T \cdot {w_j}}{\|{w_i}\|\|{w_j}\|}$. Secondly, for words with low semantic similarity but present in the target language, we employ standard label smoothing, with $\beta$ representing the smoothing parameter. Lastly, words outside the target language receive zero smoothing. Subsequently, we normalize the vector to sum up to one.
\noindent
One challenge of using this approach with current LLMs lies in the tokenization process, where words may be broken down into subwords by the tokenizer. To address this, we apply semantically aware label smoothing to the initial subword tokens. This method involves comparing the initial token with all other words in the target dataset and increasing the probability of generating similar ones. For subsequent tokens of the same word, target label smoothing is applied, which involves normal smoothing of the labels of target tokens. We use the Semantically Aware Label Smoothing (SALS) in fine-tuning our final model.
The $\hat{{\bf y_i}}$ represents the output logits corresponding to word $v_i$, and ${\bf y}_{sls}$ denotes the SLS labels. The loss component for a specific class is given by:
$$
\ell\left(\hat{\mathbf{y}}_i, \mathbf{y}_{s l s}\right)=-\frac{1}{N} \sum_{i=1}^N \mathbf{y}_{s l s} \log \left(\hat{\mathbf{y}}_i\right)
$$




\section{Experiments}

We evaluate our approach on the PHOENIX-2014T \citet{camgoz2018neural} dataset, focusing on German Sign Language videos of weather broadcasts. With $8,257$ sequences containing $1,066$ glosses and $2,887$ German words. This dataset provides a domain-specific benchmark for assessing our fine-tuned models using  BLEU score.


\noindent
\textbf{NLLB-200 } is a multilingual LLM developed by Meta \citet{costa2022no}, trained on 200 languages. Utilizing {\em SentencePiece} tokenizer~\citet{kudo2018sentencepiece} this model aims to effectively generate and process text in multiple languages, facilitating cross-lingual understanding and generation capabilities. It is trained on a vast corpus comprising 3.6B sentences from low-resource and 40.1B sentences from high-resource languages. 

\noindent
\textbf{Paraphrasing.} 
For paraphrasing English was chosen as the intermediate translation language on NLB-200 model. For each gloss-spoken pair, we translate the spoken language to English and then back to German. We use the 3.3B model with a maximum sequence length of $50$ and a beam search of $5$ for inference, we generate a total of $7040$ silver label spoken texts and add these gloss-spoken pairs to our primary training dataset.\\
\noindent
\textbf{Back Translation.} We generate synthetic glosses by switching the gloss-spoken language pairs to spoken-gloss pairs and fine-tune a model specifically for this translation task, stopping the training process after $10$ epochs. For inference, we pass the training set through the model once more, we add any generated sequences differing from the original gloss to our training set. This augmentation method results in the addition of $6523$ gloss-spoken pairs to our dataset. Consistent with the forward translation, we utilize a maximum sequence length of $100$ and a beam search of $5$ for inference. \\
\noindent
\textbf{Training.} For the Semantically Aware Label Smoothing technique (SALS) we set the cosine similarity threshold to $0.6$ to ensure that we consider only words with sufficiently high semantic similarity. We also set $\beta$ to $0.1$. For our final approach, we utilize the NLLB with 3.3B parameters. The architecture consists of 24 encoder and decoder layers. We use the AdamW optimizer \citet{loshchilov2017decoupled} to train our network with $\beta_1 = 0.9$ and $\beta_2 = 0.998$. We train the network on two NVIDIA V100 for $60$ epochs, using a maximum sequence length of $100$ and a beam search of $5$.

\noindent
\textbf{LoRA.} To further optimize model performance, we employ the LoRA \citet{hu2021lora} technique for fine-tuning. In this approach, we freeze the original model weights and train only the sign language adapter for the target task. This enables us to maintain the original functions of LLMs. Additionally, as demonstrated in Table \ref{tab:size} row $3$, LLMs with billions of parameters show the risk of overfitting. By employing the LoRA adapter, we address this concern and can leverage larger models effectively, see row $4$. Finally, the final adapter model is memory-efficient, occupying approximately $100$ Megabytes of space. For the LoRA configuration, we utilized a rank of $16$ and an alpha value of $32$.

\subsection{Results}
Table \ref{tab:res} presents a comparison between our model and previous baselines in terms of BLEU-score. The first baseline \citet{camgoz2018neural} employs an RNN encoder-decoder architecture, while the work by \citet{camgoz2020sign} utilizes a transformer encoder-decoder trained from scratch.  \citet{yin2020better} use a transformer model with FastText embeddings initialization. \citet{chen2022two} employ a pre-trained multilingual Mbart model, whereas \citet{ye2023scaling} combine the multilingual Mt5 model and GPT for translation. Our method exhibits superior performance, with $3.75\%$ relative improvement in BLEU-1 (1.98 score diff), $6.69\%$ in BLEU-4 (1.77 score diff), $5.38\%$ in ROUGE (2.78 score diff), and $2.07\%$ in CHRF++ (1.03 score diff), with significantly fewer parameters(appendix \ref{sec:appparams}), enhancing both effectiveness and efficiency. Additionally, we observe a performance discrepancy between the dev and test sets in some previous models, as seen in the first three rows. However, our method demonstrates an average score decrease of only $0.22$ across BLEU, ROUGE, and CHRF++ metrics when transitioning from the development set to the test set. Our experiments indicate that our label smoothing method contributes to improved generalization, effectively minimizing this performance gap. 

\begin{table}[t]
\centering
\begin{tabular}{lcc}
\toprule
 & BLEU-1 &  BLEU-4 \\
\midrule
600M & 52.7 & 26.5 \\
\hline
1.3B & 53.4 & 27.3 \\
\hline
3.3B & 53.1 & 27.1 \\
\hline
3.3B+LoRA & 53.8 & 27.5 \\
\bottomrule
\end{tabular}
\caption{Comparison of BLEU-score performance across different model sizes, ranging from 600M to 3.3B parameters, for gloss translation tasks.}
\label{tab:size}
\end{table}%

\noindent
\textbf{Zero Shot performance:} To further evaluate the understanding of these models for the sign language translation task, we conduct an experiment without fine-tuning the Language Models (LLMs), creating a zero-shot scenario. As shown in Table \ref{tab:res} row 6, despite the prior training of this model on the German data, the results proved to be suboptimal. This underscores the main role of fine-tuning in optimizing LLMs for {\em Gloss2Text} translation.

\noindent
\textbf{Loss Function: } To evaluate the effectiveness of our modified loss function, we conduct two set experiments. Initially, we exclude our SALS term during fine-tuning, substituting it with conventional cross-entropy loss, row 7. Subsequently, we replace the cross-entropy loss with our proposed loss for comparison, row 9. Table \ref{tab:res}, shows that our model demonstrates better performance with the integration of semantically aware label smoothing (see rows 7 and 9). The NLLB-SALSloss system demonstrates an average improvement of $1.68$ points on the development set and $1.06$ points on the test set over the NLLB-FineTuned system across BLEU, ROUGE, and CHRF++ metrics.

\commentOUT{
\noindent
\textbf{Back Translation model} The quality of the back-translated model directly correlates with the quality of the generated data and, consequently, influences the final performance. Table \ref{tab:back} presents a comparison between two models, both trained with the same configuration. As depicted, the NLLB-200 model also surpasses the Mbart-50 in the back translation task. Through fine-tuning, the NLLB-200 model generates higher-quality pseudo-parallel data, as shown by better BLEU scores on the validation set of the text-to-gloss translation.
}

\noindent
\textbf{Data Augmentation Techniques: } 
We also compare our method with various data augmentations, namely paraphrasing and backward translation, for the gloss translation task. Table \ref{tab:res}, row 8, demonstrates the improvements achieved by applying these augmentations using the cross-entropy loss.

\noindent
\textbf{Model Size: }
We utilize NLLB-200 models ranging from 600M to 3.3B parameters. As illustrated in Table \ref{tab:size}, larger models generally lead to better translation performance. However, with the largest model, we observed signs of overfitting to the dataset. The reason is the fine-tuning dataset is relatively small compared to the number of parameters in the model. To address this issue, we explore LoRA techniques to enhance model performance and mitigate overfitting.

\section{Conclusion}
We explored a comprehensive exploration of {\em Gloss2Txt} translation for sign language using large language models and the PHOENIX-2014T dataset. We evaluated different model architectures, data augmentations, and loss functions. Our experiments showed that our Semantically Aware Label Smoothing technique significantly improves translation quality over state-of-the-art models.

\section*{Limitation and Open problems}
While glosses provide additional structured annotation of sign language videos, they do not fully capture the complexity of sign language communication. Facial expressions, which are part of conveying meaning in sign language, are often not represented in gloss annotations. Additionally, gestures involving pointing to specific locations or objects are typically omitted in gloss representations. Even with the visual modality, such expressive nuances are often absent in text gloss representations, but are captured in the corresponding spoken language translations.


Also, it's important to acknowledge that datasets used in sign language often have a domain-specific vocabulary. This vocabulary may not always reflect the everyday activities and interactions prevalent in the deaf community. This potentially limits the scope and applicability of sign language systems developed using such datasets and calls for additional benchmarks and evaluation methodologies for sign language translation. 

\section*{Acknowledgements}
This work was partially supported by the \mbox{Amazon Research Award} and resources provided by the Office of Research Computing at George Mason University (URL: \url{https://orc.gmu.edu}) and funded in part by grants from the National Science Foundation (Award Number 2018631).
Antonios Anastasopoulos is additionally generously supported by the National Science Foundation under grant IIS-2327143.

\bibliography{acl_latex}

\clearpage

\appendix
\section{Trainable parameters}
\label{sec:appparams}
In comparison with previous SOTA methods, we also investigate the total number of trainable parameters. Although our method uses a larger model, it does not require many parameters to train due to the use of LoRA adapters. This allows the original LLM to be used for other tasks, with the adapter applied only when sign language gloss translation is needed. In our experiments, all the linear layers are used in the LoRA fine-tuning for optimal performance. Table \ref{tab:params} compares the trainable parameters.

\begin{table}[ht]
\centering
\begin{tabular}[t]{lc}
\toprule
& Trainable Params \\
\midrule
\citet{camgoz2018neural}  & 100M  \\
\citet{camgoz2020sign} & 150M   \\
\citet{yin2020better} &  130M \\
\citet{chen2022two} & 611M \\
\citet{ye2023scaling} &  570M \\
\hline
NLLB-200 & 26M \\
\bottomrule
\end{tabular}
\caption{Comparison of model sizes and the number of trainable parameters for various methods, highlighting the efficiency of our approach with LoRA adapters.}
\label{tab:params}
\end{table}%


\section{Architecture Type}
\label{sec:aptype}
We initially investigated various architectures for the task of gloss translation. To our knowledge, this is the first attempt to compare the impact of Large Language Model LLM architectures on gloss translation. We explore both decoder-only models, specifically GPT-2, and encoder-decoder models, including MT5-Base, Mbart-50, and NLLB-200. The results presented in Table \ref{tab:archs} illustrate their performance. We use the NLLB with 600M parameters to closely match the others in terms of trainable parameters. Our experiments show that encoder-decoder models perform better than the GPT model. This could potentially be attributed to the diverse pretraining datasets utilized by encoder-decoder models, enabling them to comprehend low-resource tasks more effectively and leverage their multilingual capabilities.

\begin{table}[h]
\centering
\begin{tabular}[t]{lccc}
\toprule
& \#Params & BLEU-1 &  BLEU-4  \\
\midrule
GPT-2 Scratch & 124M & 45.21 & 19.3\\
\hline
GPT-2(German) & 137M & 49.4 & 24.94 \\
\hline
M-Bart-25 & 610M & 52.1 & 26.4 \\
\hline
MT5-Base & 580M  & 50.1 & 24.4 \\
\hline
NLLB-200 & 600M & 52.7 & 26.5 \\
\bottomrule
\end{tabular}
\caption{Comparison of performance across different architecture types, including encoder-decoder and decoder-only models, for gloss translation tasks. For a fair comparison, pre-trained models were selected to have approximately the same number of parameters across all architecture types.}
\label{tab:archs}
\end{table}%

\begin{figure}[h]
\centerline{\includegraphics[width=1\linewidth]{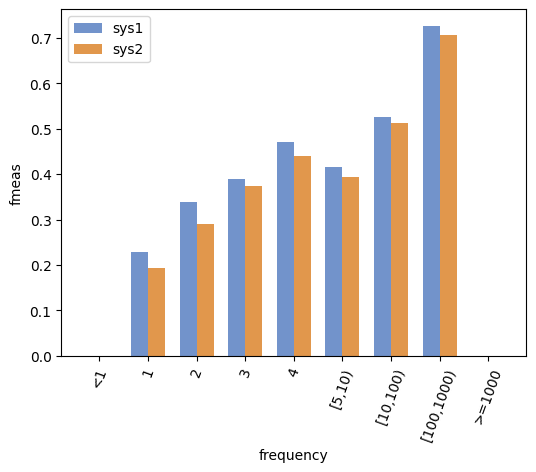}}
\caption{Comparison of word-level F-measure scores across different frequency buckets with \citet{chen2022two}.}
\label{fig:abword}
\end{figure}

\begin{table}[h]
\centering
\begin{tabular}[t]{lcc}
\toprule
& BLEU-1 &  BLEU-4  \\
\midrule
Nbart-50 & 64.57 & 25.48 \\
\hline
NLLB-200 Aug & 67.63 & 26.47 \\
\bottomrule
\end{tabular}
\caption{Back translation performance under various fine-tuning approaches. NLLB-200 archives better performance in generating pseudo-gloss annotations for the text-to-gloss Dev set.}
\label{tab:back}
\end{table}%

\section{Back Translation model} The quality of the back-translated model directly correlates with the quality of the generated data and, consequently, influences the final performance. Table \ref{tab:back} presents a comparison between two models, both trained with the same configuration. As depicted, the NLLB-200 model also surpasses the Mbart-25 in the back translation task. Through fine-tuning, the NLLB-200 model generates higher-quality pseudo-parallel data, as shown by better BLEU scores on the validation set of the text-to-gloss translation.

\section{Analysis with previous SOTA}
We compare our results with previous state-of-the-art methods. Our initial comparison focuses on the length of the predictions. It appears that our method generates shorter sentences compared to the more lengthy sentences produced by the previous state-of-the-art. Notably, both methods used the same hyperparameters for generation: a length penalty of $1$, a maximum length of $100$, and a beam search size of $5$. Our method's generation ratio is $0.9808 (ref=8458, out=8296)$, whereas the ratio for \citet{chen2022two} is $1.0233 (ref=8458, out=8655)$.

\begin{figure}[t]
\centerline{\includegraphics[width=1\linewidth]{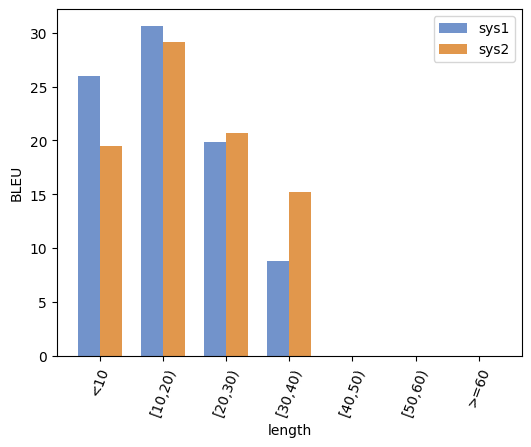}}
\caption{Comparison of sentence-level F-measure scores across different frequency buckets with \citet{chen2022two}.}
\label{fig:absent}
\end{figure}

\textbf{Word Accuracies:} We also evaluated word-level F-measure scores across different frequency buckets. This analysis allows us to observe the performance of our method relative to the frequency of the words in the dataset. Our method shows consistent improvements across most frequency buckets compared to the previous state-of-the-art. This suggests that our approach is more effective at predicting both common and rare words accurately. Detailed results are shown in Table \ref{fig:abword}. We further analyzed the performance of our method by comparing sentence-level F-measure scores across different sentence length buckets. The results indicate that our method excels in predicting shorter sentences but lags behind in longer sentences, likely due to the other method generating lengthier predictions. These findings are illustrated in Figure \ref{fig:absent}. 

\textbf{Translation examples:} Tables \ref{table:example_translationsours} and \ref{table:example_translationsthem} present several translation examples along with their BLEU scores. The first table highlights instances where our translations outperform, while the second table showcases failure cases where the previous state-of-the-art methods perform better.

\pagebreak

\newcolumntype{L}[1]{>{\raggedright\let\newline\\\arraybackslash\hspace{3pt}}m{#1}}

\begin{table*}[h!]
\centering
	\begin{tabular}{ c  L{\dimexpr.8\textwidth}  c }
\noalign{\vskip 2mm}
\arrayrulecolor{black} 
        \hline\hline
        Reference & später ist es meist trocken. & BLEU \\
        \arrayrulecolor{gray!50} 
        \hline
        \citet{chen2022two} &  später wird es aber schon wieder trockener. &  18.27 \\
        \arrayrulecolor{gray!50} 
        \hline
        Ours & später ist es meist trocken. & 100 \\
        \arrayrulecolor{black} 
        \hline\hline
        Reference & dort morgen bis zweiundzwanzig grad. &  \\
        \arrayrulecolor{gray!50} 
        \hline
        \citet{chen2022two} & morgen temperaturen von zweiundzwanzig grad im breisgau bis zweiundzwanzig grad am oberrhein. &  19.14 \\
        \arrayrulecolor{gray!50} 
        \hline
        Ours & dort morgen bis zweiundzwanzig grad. & 100 \\
        \arrayrulecolor{black} 
        \hline\hline
        Reference &  der deutsche wetterdienst hat entsprechende warnungen herausgegeben. &  \\
        \arrayrulecolor{gray!50} 
        \hline
        \citet{chen2022two} &   es gelten entsprechende warnungen des deutschen wetterdienstes. &  21.73 \\
        \arrayrulecolor{gray!50} 
        \hline
        Ours & der deutsche wetterdienst hat entsprechende warnungen herausgegeben. & 100 \\
        \arrayrulecolor{black} 
        \hline\hline
        Reference &  jetzt wünsche ich ihnen noch einen schönen abend. &  \\
        \arrayrulecolor{gray!50} 
        \hline
        \citet{chen2022two} &   guten abend liebe zuschauer. &  12.83 \\
        \arrayrulecolor{gray!50} 
        \hline
        Ours &  und jetzt wünsche ich ihnen noch einen schönen abend. & 89.09 \\
        \arrayrulecolor{black} 
        \hline\hline
        Reference &  auf den bergen sind orkanartige böen möglich. &  \\
        \arrayrulecolor{gray!50} 
        \hline
        \citet{chen2022two} &   im bergland sind zum teil orkanartige böen möglich. &  40.32 \\
        \arrayrulecolor{gray!50} 
        \hline
        Ours & auf den bergen sind orkanartige böen möglich. & 100 \\
        \arrayrulecolor{black} 
        \hline\hline
        Reference & am montag meist trocken bei einer mischung aus sonne und wolken. &  \\
        \arrayrulecolor{gray!50} 
        \hline
        \citet{chen2022two} &    am montag ist es meist trocken sonne und wolken gibt es eine mischung aus nebel und sonne. &  19.20 \\
        \arrayrulecolor{gray!50} 
        \hline
        Ours &  am montag bleibt es meist trocken bei einer mischung aus sonne und wolken. & 74.66 \\
        \arrayrulecolor{black} 
        \hline\hline
        Reference &  der deutsche wetterdienst hat entsprechende unwetterwarnungen herausgegeben. &  \\
        \arrayrulecolor{gray!50} 
        \hline
        \citet{chen2022two} &  es gelten entsprechende warnungen des deutschen wetterdienstes. &  16.51 \\
        \arrayrulecolor{gray!50} 
        \hline
        Ours &  der deutsche wetterdienst hat entsprechende warnungen herausgegeben. & 65.80 \\
        \arrayrulecolor{black} 
        \hline\hline
        Reference &  ähnliches wetter auch am donnerstag. &  \\
        \arrayrulecolor{gray!50} 
        \hline
        \citet{chen2022two} &   und nun die wettervorhersage für morgen donnerstag den achten juli. &  11.64 \\
        \arrayrulecolor{gray!50} 
        \hline
        Ours &  ähnliches wetter dann auch am donnerstag. & 59.15 \\
        \arrayrulecolor{black} 
        \hline\hline
        Reference & jetzt wünsche ich ihnen noch einen schönen abend. &  \\
        \arrayrulecolor{gray!50} 
        \hline
        \citet{chen2022two} &   ihnen noch einen schönen abend und machen sie es gut. &  42.64 \\
        \arrayrulecolor{gray!50} 
        \hline
        Ours &  und jetzt wünsche ich ihnen noch einen schönen abend. & 89.09 \\
        \arrayrulecolor{black} 
        \hline\hline
        Reference &  heute nacht liegen die werte zwischen vierzehn und sieben grad. &  \\
        \arrayrulecolor{gray!50} 
        \hline
        \citet{chen2022two} &   heute nacht vierzehn bis sieben grad. &  24.00 \\
        \arrayrulecolor{gray!50} 
        \hline
        Ours &  heute nacht werte zwischen vierzehn und sieben grad. & 66.51 \\
        \arrayrulecolor{black} 
        \hline\hline

\hline 
\end{tabular}
\caption{Here are example translations comparing our method with the previous state of the art \cite{chen2022two}. These examples, provided by \cite{compare-mt}, highlight instances where our method achieves a higher BLEU score.}
\label{table:example_translationsours}
\end{table*}

\pagebreak

\newcolumntype{L}[1]{>{\raggedright\let\newline\\\arraybackslash\hspace{3pt}}m{#1}}

\begin{table*}[h!]
\centering
	\begin{tabular}{ c  L{\dimexpr.8\textwidth}  c }
\noalign{\vskip 2mm}
\arrayrulecolor{black} 
        \hline\hline
        Reference & auch am tag wieder viel sonnenschein später bilden sich hier und da ein paar quellwolken. & BLEU \\
        \arrayrulecolor{gray!50} 
        \hline
        \citet{chen2022two} &  auch am tag viel sonne später hier und da ein paar quellwolken. &  50.93 \\
        \arrayrulecolor{gray!50} 
        \hline
        Ours & auch am tag scheint verbreitet die sonne später kommen an den küsten wieder dichtere wolken auf. & 15.09 \\
        \arrayrulecolor{black} 
        \hline\hline
        Reference & sonst ist es recht freundlich. &  \\
        \arrayrulecolor{gray!50} 
        \hline
        \citet{chen2022two} & sonst ist es recht freundlich. &  100 \\
        \arrayrulecolor{gray!50} 
        \hline
        Ours & ansonsten wird es recht freundlich. & 60.42 \\
        \arrayrulecolor{black} 
        \hline\hline
        Reference &  im südosten regnet es teilweise länger. &  \\
        \arrayrulecolor{gray!50} 
        \hline
        \citet{chen2022two} &   im südosten regnet es teilweise ergiebig. &  70.34 \\
        \arrayrulecolor{gray!50} 
        \hline
        Ours & in der südosthälfte regnet es teilweise ergiebig. & 30.73 \\
        \arrayrulecolor{black} 
        \hline\hline
        Reference &  im süden bleibt es morgen unter hochdruckeinfluss zunächst noch recht freundlich und warm. &  \\
        \arrayrulecolor{gray!50} 
        \hline
        \citet{chen2022two} &   im süden deutschlands bleibt es morgen unter hochdruckeinfluss noch weitgehend freundlich und warm. &  53.95 \\
        \arrayrulecolor{gray!50} 
        \hline
        Ours &  in der südhälfte bestimmt hochdruckeinfluss morgen unser wetter und es bleibt noch ziemlich warm. & 14.05 \\
        \arrayrulecolor{black} 
        \hline\hline
        Reference &  morgen reichen die temperaturen von einem grad im vogtland bis neun grad am oberrhein. &  \\
        \arrayrulecolor{gray!50} 
        \hline
        \citet{chen2022two} & morgen reichen die temperaturen von einem grad im vogtland bis neun grad am oberrhein. &  100 \\
        \arrayrulecolor{gray!50} 
        \hline
        Ours & morgen temperaturen im vogtland bis neun grad. & 27.09 \\
        \arrayrulecolor{black} 
        \hline\hline
        Reference &  auch am tag wieder viel sonnenschein später bilden sich hier und da ein paar quellwolken. &  \\
        \arrayrulecolor{gray!50} 
        \hline
        \citet{chen2022two} &  auch am tag viel sonne später hier und da ein paar quellwolken. &  50.93 \\
        \arrayrulecolor{gray!50} 
        \hline
        Ours &  auch am tag scheint verbreitet die sonne später kommen an den küsten wieder dichtere wolken auf. & 15.09 \\
        \arrayrulecolor{black} 
        \hline\hline
        Reference &  morgen vormittag an der ostsee noch starke böen sonst weht der wind schwach bis mäßig aus ost bis südost. &  \\
        \arrayrulecolor{gray!50} 
        \hline
        \citet{chen2022two} &   morgen vormittag an der nordsee noch kräftige böen sonst weht der wind schwach bis mäßig. &  50.76 \\
        \arrayrulecolor{gray!50} 
        \hline
        Ours &  morgen vormittags an der nordsee starke bis stürmische böen sonst meist nur schwacher bis mäßiger wind aus süd bis südwest. & 11.84 \\
        \arrayrulecolor{black} 
        \hline\hline
        Reference & morgen muss verbreitet mit teilweise kräftigen schauern und gewittern gerechnet werden. &  \\
        \arrayrulecolor{gray!50} 
        \hline
        \citet{chen2022two} &   morgen muss mit teilweise unwetterartigen schauern und gewittern gerechnet werden. &  56.53 \\
        \arrayrulecolor{gray!50} 
        \hline
        Ours &  morgen gibt es dort zum teil kräftige schauer und gewitter. & 11.76 \\
        \arrayrulecolor{black} 
        \hline\hline
        Reference &  sonst viel sonnenschein. &  \\
        \arrayrulecolor{gray!50} 
        \hline
        \citet{chen2022two} &   sonst viel sonnenschein. &  100 \\
        \arrayrulecolor{gray!50} 
        \hline
        Ours &  ansonsten scheint verbreitet die sonne. & 19.30 \\
        \arrayrulecolor{black} 
        \hline\hline

\hline 
\end{tabular}
\caption{Here are example translations of failed cases where our method obtains a lower BLEU score.}
\label{table:example_translationsthem}
\end{table*}

\end{document}